\documentclass[master, final, oneside]{KUThesis}
\usepackage{multirow}

\usepackage{array}
\newcommand{\PreserveBackslash}[1]{\let\temp=\\#1\let\\=\temp}
\newcolumntype{C}[1]{>{\PreserveBackslash\centering}p{#1}}
\newcolumntype{R}[1]{>{\PreserveBackslash\raggedleft}p{#1}}
\newcolumntype{L}[1]{>{\PreserveBackslash\raggedright}p{#1}}
\usepackage{float} 
\usepackage{enumitem}
\usepackage{subfigure}

\title{3}{
Improving Object Detection, Multi-object Tracking, and Re-Identification for Disaster Response Drones
}{}

\author[korean]{}{}{}
\author[chinese]{}{}{}
\author[english]{}{}{}
\advisor{}{}{}
\department{}
\graduateDate{}{}
\submitDate{}{}{}
\approvalDate{}{}{}

\referee[1]{}            
\referee[2]{}
\referee[3]{}
\referee[4]{}
\referee[5]{}

\captionLineSpacing{150}
\abstractLineSpacing{200}
\krAbstractLineSpacing{200}
\TOCLineSpacing{200}
\contentLineSpacing{200}
\acknowledgementLineSpacing{200}

\begin{document}
\chapter{Introduction}

Object tracking has long been an interesting field in computer vision. The main purpose of tracking is to recognize an object and follow its trajectory, which is used as an important solution in video analysis, action recognition, smart elderly care, and human-computer interaction \cite{zhang2021fairmot, xiao2017joint}.

The tracking system is able to expand by utilizing spatial information through multiple cameras or moving camera devices like drones. These technologies that automatically detect and track objects or people using multiple cameras are widely used for unmanned surveillance systems of drones or vehicles, analysis of sports games, crime prevention, and manufacturing systems.  To cover all situations, it is important to implement such a system through multiple cameras. Re-identification is a fundamental process for a multi-camera system to determine the identity of objects by recognizing information from each different non-overlapping camera, which shoots different spaces without any information about a timeline or the same space, but completely different situations are given.

\begin{figure}[htbp]
\centering
    \includegraphics[width=1.0\textwidth]{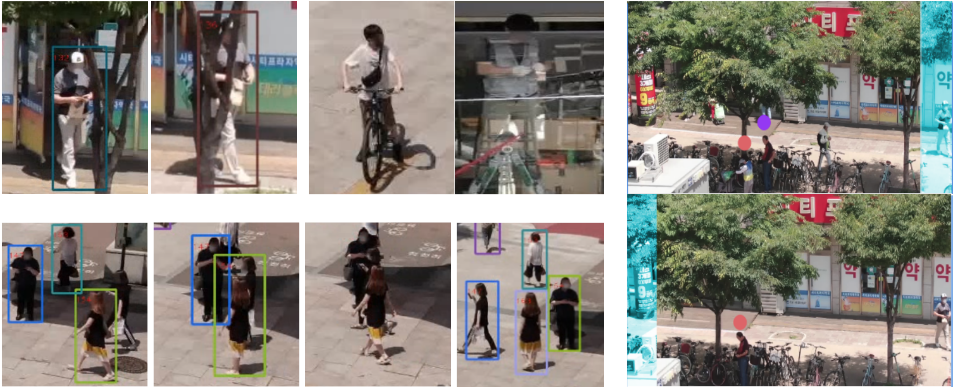}
    \caption{Challenges. (top-left) occlusion and detection errors.  (bottom-left) fragmentation. (right) global motion.}
    \label{challenages}
\end{figure}
   
There are lots of challenges in object tracking and re-identification, such as tracking issues like ID switches, fragmentation, and an external disturbance like the global motion of cameras in the multiple object tracking and ID associations. Detector errors even occur like recognizing the motorcycle as a human instead of a person who rides a bicycle, as shown in Figure \ref{challenages}. 
It may also lose track of the human because of reflection on windows or occlusion due to others. 
Furthermore, different scales or multiple views from different cameras can make it difficult to properly identify who is who, and fail re-identification. In addition, real-time processing might be required in various environments with insufficient resources, efficient deep learning techniques could be applied as well.

Our purpose is to improve these practical difficulties. So, we propose a simple multi-camera system, which combines two steps, the tracking of the unique persons in a single camera and data association of tracklets from different cameras. Our approaches demonstrate resolving two types of situations; one is crowded scenes where occlusions generally happen, and ID fragmentation is caused by many fast-moving people, and the other is a real disaster situation with irregular camera movements. We implement a multi-camera system for drones with FairMOT \cite{zhang2021fairmot} and OSNet \cite{9011001} for the first challenge which requires fast processing. To handle the second situation, YOLOv5 and DeepSORT \cite{wojke2017simple} are applied to disaster response drone systems as a more accurate detector and tracker. OSNet \cite{9011001} is used in both cases for extracting better representational feature vectors for tracklet association.

The accuracy of our first approach (85.71\%) is slightly improved compared to our baseline, FairMOT (85.44\%) in the validation dataset. In the final results calculated based on L2-norm error, the baseline was 48.1, while the proposed model combination was 34.9, which is a great reduction of error by a margin of 27.4\%. In the second approach, although DeepSORT only processes a quarter of all frames due to hardware and time limitations, our model with DeepSORT (42.9\%) outperforms FairMOT (71.4\%) in terms of recall.

In summary, the contributions of this work include the following:

\begin{itemize}[noitemsep,nolistsep]
\item We propose a fast and lightweight multi-camera system that handles detection and tracking issues such as occlusion, ID fragmentation, and global motion through the addition of re-identification and tracklet association steps in crowded scenes with many fast-moving objects.
\item We present both quantitative and qualitative evaluations of our approach, focusing on enhancing object detection performance and the post-processing after the re-ID block, regarding the difficulties of irregular movement of the camera and disturbances in the background.
\item We resolve constraints on time and hardware resources corresponding to the actual disaster situation with the implementation techniques.
\item We demonstrate that our approaches are suitable combinations of object trackers and re-ID models by winning second\footnote{2020 AI Grand Challenge result - https://www.ai-challenge.kr/sub03/view/id/557} and third\footnote{2021 AI Grand Challenge news - https://n.news.naver.com/article/003/0010879643} place in the ``2020  and 2021 AI Grand Challenge for Disaster Response Drones'' organized by the Korean Ministry of Science and ICT, respectively.
\end{itemize}
\chapter{Related Work}

\textbf{Tracking by Detection} \cite{bewley2016simple, wojke2017simple}. 
Most object trackers were affected by Tracking by Detection. It consists of an object detection part and an identity association part, which is namely the two-stage object tracker. SORT \cite{bewley2016simple} used the Kalman filter for the first time to predict the next location of tracklets, distinguishing tracklets with Hungarian algorithm. SORT is simple and imposes little burden in terms of computation. This approach achieves a tracking speed of 260 Hz \cite{bewley2016simple}. 

DeepSORT \cite{wojke2017simple} reduces ID switching issues of SORT. The Kalman filter was not enough to solve the ID switches caused by occlusion. Thus, appearance descriptors from feature extraction are used to match detection information and tracks. Kalman filter predicts tracklet locations in the following frames and computes the Mahalanobis distance between the predicted and detected boxes. 

Recently, lots of trackers are based on DeepSORT with some modification for feature processing \cite{Chen_2021_ICCV}. YOLOv5\footnote{YOLOv5 link - https://github.com/ultralytics/yolov5} is frequently employed as the backbone network to improve detection performance.

\textbf{Joint Detection and Tracking} \cite{xiao2017joint, wang2020towards, zhang2021fairmot, zhou2020tracking, Lu_2020_CVPR, Wu_2020_CVPR, wu2021track}. Two-stage object trackers have a strong dependency on the performance of detectors. Generally, they are more accurate but slower than one-shot object trackers. On the other hand, one-shot trackers generally share a backbone network, which makes it faster and lighter compared to two-stage object trackers.

JDE \cite{wang2020towards} introduces a shared model, which builds on detection elements such as bounding boxes and appearance features for re-ID. It attains a real-time processing speed, 22 to 40 FPS depending on the input resolution. FairMOT \cite{zhang2021fairmot} is built on JDE with CenterNet \cite{zhou2019objects}. FairMOT mostly follows JDE's paradigm. However, it overcame the weaknesses of JDE by several implementation techniques. Anchor-free detection makes it get better re-ID features, and multi-layer feature aggregation improves its scaling performance. Also, low dimensional features help reduce over-fitting problems in tracking.

TraDes \cite{wu2021track} tried to solve occlusion issues, maintaining the concept of the one-shot object tracker, using CVA (Cost Volume-based Association) and MFW (Motion-guided Feature Warper). Both modules learn the relation of previous tracklets and current ones by coordinating the tracking and detection modules, without conflicts of each loss function.

\textbf{Person Re-Identification} \cite{9011001, ristani2016performance}.
Person re-identification (re-ID) is a fundamental task in distributed multi-camera surveillance, which aims to match people appearing in different non-overlapping camera views. As an instance-level recognition problem, person re-ID faces two major challenges. First, the intra-class (instance/identity) variations are typically big due to the changes in camera viewing conditions. Second, there are also small inter-class variations. The key to overcoming these two challenges is to learn scale-invariant feature vectors from cropped images \cite{9011001}. The representational features will be used to compute distances between different image snippets, which serve as a ground for re-ID.

Duke Multi-Target Multi-Camera Tracking (MTMCT) \cite{ristani2016performance} introduced the dataset for multi Camera person tracking, which consists of 2834 IDs, 85 minute-long with 8 camera view videos. A usual way to track a person in multi-camera was to track and associate with appearance and locations \cite{ristani2018features}. The performance of the model was determined by the position of every person at all times from video streams taken from multiple cameras.

\chapter{Object Tracking}

Object tracking integrates detection and tracking into a single process. 
Detection localizes the new object or person while tracking determines whether a person or object is the same. Tracking typically obtains a bounding box and centroid and finds a new centroid using measurements such as Euclidean distance. It associates IDs with short distances and a new ID is assigned to the object which does not belong to the group \cite{bewley2016simple}. Basically, the former is computationally more expensive than the latter.

\begin{figure}[ht]
\centering
    \includegraphics[width=1.0\textwidth]{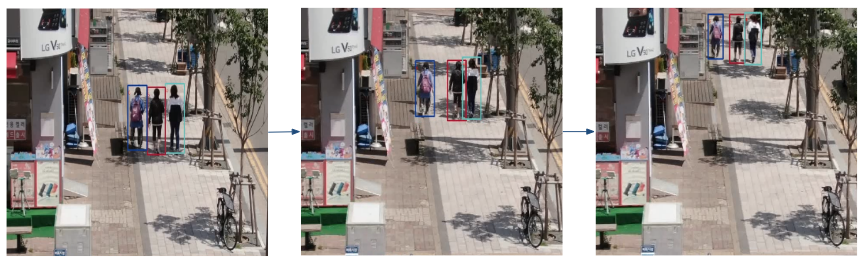}
    \caption{Object tracking is a process of locating and identifying a moving object over time using a camera.}
    \label{tracking}
\end{figure}

Object tracking has two categories: Tracking By Detection \cite{bewley2016simple, wojke2017simple} and Joint Tracking and Detection \cite{xiao2017joint, wang2020towards, zhang2021fairmot, zhou2020tracking, Lu_2020_CVPR, Wu_2020_CVPR}. Tracking By Detection is called the two-step model, which connects features extracted from each bounding box with tracklets after detection object localization. Joint Tracking and Detection using a single model reduces inference time by re-using backbone features for re-ID, but performance drops compared to the two-step models.

\textbf{FairMOT} \cite{zhang2021fairmot}. 
We briefly review one of the fast object trackers we use, which is a one-shot tracker, but gets better performance than JDE \cite{wang2020towards}. FairMOT has a similar architecture to JDE. 
However, it uses CenterNet \cite{zhou2019objects} which stacks multiple scales of the hourglass network. The hourglass structure aggregates each different scaled feature with an auto-encoder instead of a Feature Pyramid Network \cite{Lin_2017_CVPR}, which increases accuracy. Consequently, general and detailed information is combined through down-scaled features and up-scaled features. 
It alleviates the ID switching problem and improves the scaling performances of tracking. Furthermore, CenterNet resolves detection errors caused by ambiguities of the anchor by making the object tracker get proper feature embedding of re-ID. From these things, we may assume that FairMOT is able to track the densely distributed objects well in crowded scenes.

\textbf{DeepSORT} \cite{wojke2017simple}.
DeepSORT suggests a process to link bounding boxes and tracks with a simple Kalman filter and objects' appearance descriptor. The Kalman filter predicts short-term movements of objects in the image space, which can only provide a rough estimation of object location \cite{bewley2016simple}. 
ID fragmentation, occlusion, and different views make tracking capricious. DeepSORT improves the long-term trajectory by using a cost matrix of the deep appearance descriptor. Despite recent advances in one-shot object trackers\cite{zhang2021fairmot, zhou2020tracking, Lu_2020_CVPR, Wu_2020_CVPR}, the DeepSORT algorithm is still employed. As introduced in \cite{Chen_2021_ICCV}, many two-step model derivatives are built on DeepSORT. One of the advantages of two-step object trackers is their flexibility. Most methods that separate detection and tracking attach suitable models for each task. They crop the Regions of Interest (ROI) with bounding boxes using detectors, and then extract re-ID feature embeddings.  It is very useful to implement tracking depending on the situation. So, DeepSORT is a good example of our camera system with regard to these points.

\chapter{Multi-Object Tracking in Multi-Camera Systems}

\begin{figure}[ht] 
\centering
    \includegraphics[width=1.0\textwidth]{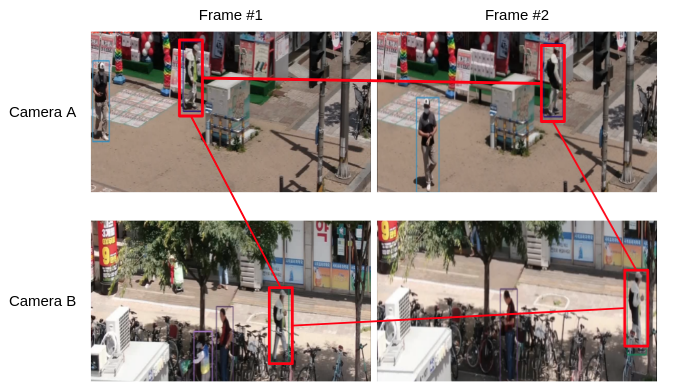}
    \caption{Person re-identification with multiple cameras.}
    \label{reid}
\end{figure}

Person Re-Identification is a fundamental task in multiple camera systems in Figure \ref{reid}, which is to find a person of interest among many people in different cameras \cite{zheng2016person}. 
It is important to embed features from the image to distinguish a specific person. Although the re-identification module is already in the object tracking stage, we further take advantage of an effective re-ID model to ensemble the result.

OSNet \cite{9011001} is the re-ID feature extractor that we employ for the implementation. When it comes to constructing a multi-camera system, efficiency is one of the important factors. As one good solution, OSNet designs its network with depth-wise separable convolutions. We believe that similarity-based re-identification is essential for developing advanced MOT algorithms.

\begin{figure}[ht]
\centering
    \includegraphics[width=1.0\textwidth]{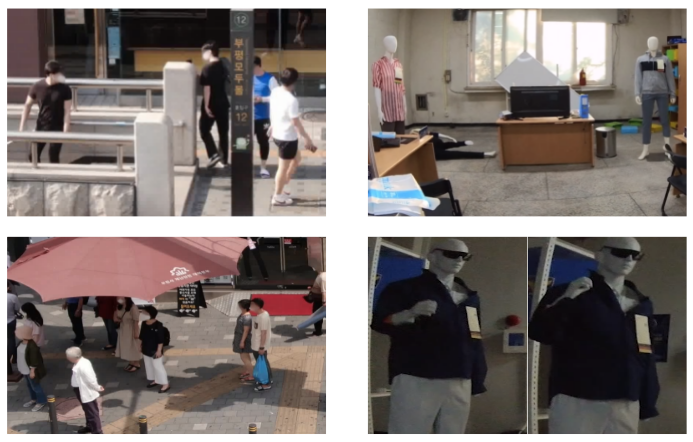}
    \caption{Datasets. (left column) Counting moving objects.  (right column) Objects in moving cameras.}
    \label{compare_dataset}
\end{figure}

Figure \ref{compare_dataset} shows two cases; one is counting the person with movements in three different cameras, and the other is to determine whether a person filmed on multiple moving cameras (e.g. drones) is moving parts of his/her body. Proper selection of the tracker is one of the important points in re-identification. Depending on the conditions on which object tracking is performed, the adequate choice of the tracking model may have a profound effect on performance.

\section{Study-1: Counting the Unique Persons with Movements}

\subsection{Problem Definition} 
The goal of Study-1 is to find the unique persons in multi-camera videos which has various camera angles. We assume a densely crowded scene where there are movements of people without much camera motion. The expected problems are ID fragmentation caused by fast-moving people or people going out of the scene and occlusions. Congested people cause overlapping between people, which makes confusion in finding the exact bounding boxes. Additionally, the inference time to track the number of people is one of the issues to be addressed.

Object trackers need to be faster but with minimal performance degradation, while mitigating all these limitations. Single-shot tracker is an essential choice \cite{zhang2021fairmot}. Lighter and faster models are required as a solution for re-ID \cite{9011001} to decrease latency.

\subsection{Architecture Overview}
Our architecture is a combination of the FairMOT \cite{zhang2021fairmot} Tracker and the OSNet Feature extractor \cite{9011001} with greedy tracklet association. 

FairMOT \cite{zhang2021fairmot} employs CenterNet \cite{zhou2019objects} as the detector, which is fast, anchor-free, and improves multi-scale feature using DLA (Deep Layer Aggregation) \cite{yu2018deep} backbone network. This indicates that FairMOT is a suitable choice for crowded images or tracking fast-moving objects to find the centroid of objects. 

OSNet \cite{9011001} is also specialized for multi-scale predictions, which are trained to be robust to different scale objects and multiple views from heterogeneous cameras. However, the training dataset for OSNet is currently not available at hand, so we have no choice but to use pre-trained weights.

Furthermore, these two combined models are lightweight. FairMOT \cite{zhang2021fairmot} is a fast single-shot object tracker, and OSNet \cite{9011001}, which consists of depth-wise separable convolution networks, is an efficient re-ID model. So, we considered using a combination of the two models to address our concerns about inference time.

\begin{figure}[ht]
\centering
    \includegraphics[width=1.0\textwidth]{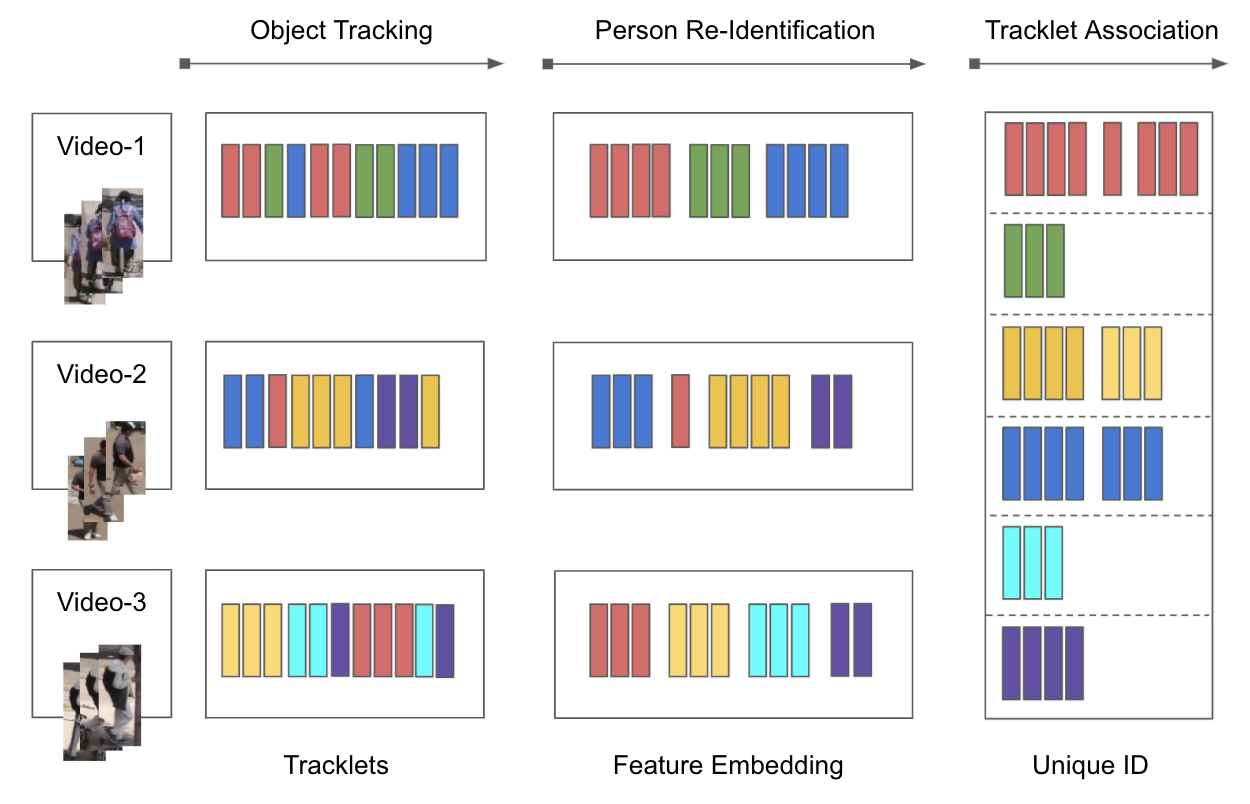}
    \caption{Data flow. First, FairMOT tracks the person in each video. Second, representational features are extracted from OSNet. Third, tracklets are associated with other tracklets extracted from different cameras.}
    \label{dataflow}
\end{figure}

The overall computation process of Figure \ref{dataflow} is described as follows.
First, the object tracker tracks the person by each camera. These tracklets move to the next stage. However, some of tracklets processed from the input images may cause ID switching due to global motion, occlusion, fragmentation, etc. These problems are supplemented in the next stage.

Second, with the tracklets from the object tracker, representational features are extracted from OSNet. The feature extractor makes each tracklet embedded as a 512-dimensional vector. Initially, we consider using the FairMOT feature extractor without OSNet to shorten inference time. However, considering the performance gain, the reduced inference time by ablating the OSNet re-ID feature extractor is trivial. The details are discussed in the next section.

Thirdly, intra-camera and inter-camera tracklet association is performed, i.e., re-ID is first performed on tracklets extracted within cameras, and then the tracklets are associated with other tracklets extracted from different cameras. We used L2 distance to measure the difference between tracklets and threshold the distance between tracklets. The measured information is classified into each unique ID.

\begin{figure}[ht]
\centering
    \includegraphics[width=1.0\textwidth]{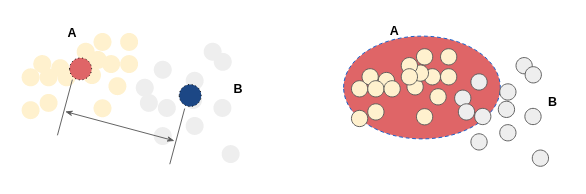}
    \caption{Methodology. Euclidean distance vs. voting method, (left) The Euclidean method is a simple distance between the mean of re-ID features. (right) If more than half of cluster A is inside cluster B, then merge A and B.
}
    \label{association}
\end{figure}

As for the methodology in the association stage, the Euclidean method and the Voting method are reviewed as in Figure \ref{association}. In the Euclidean method, based on the average value of each embedding value, e.g., Person-A and Person-B, the Euclidean distance with other IDs was measured and classified on a threshold basis. In the Voting method, if more than half of the Person-B is inside the Person-A cluster, the two clusters of Person-A and Person-B are merged. The Voting method showed more accurate performance in the small validation dataset, while some performance degradation occurred for large-scale test datasets. The cause is assumed that errors occur in a large number of test datasets for the naive Euclidean method, due to the distribution of small inter-class and relatively large intra-class according to the distribution of data.

\subsection{Experiment Setup}

\begin{figure}[ht]
\centering
    \includegraphics[width=1.0\textwidth]{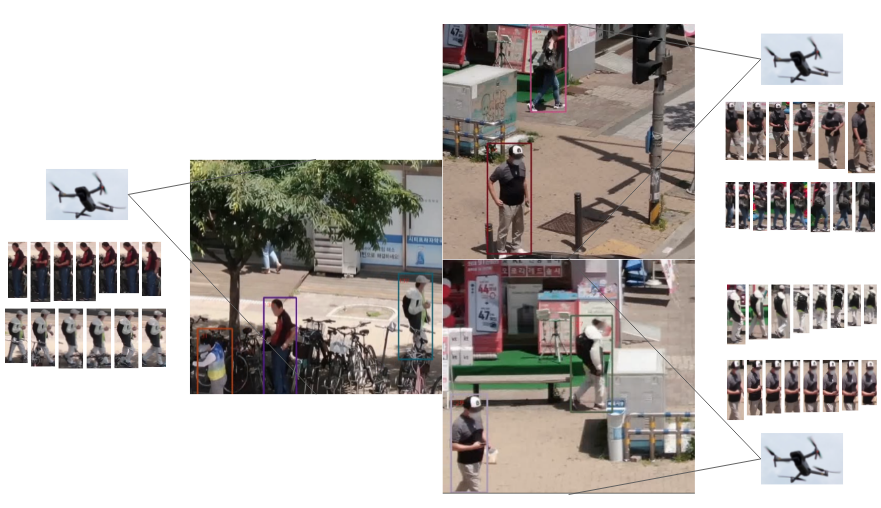}
    \caption{Validation Dataset. Fixed cameras from multiple views.}
    \label{multiple_camera}
\end{figure}

\textbf{Dataset}. CrowdedHuman and MOT17 data are the training datasets for object tracking. 
However, there is no other way because most of the MTMCT (Multi-Target Multi-Camera Tracking) datasets were banned because of privacy issues. So only pre-trained data from market1501 was allowed in re-identification. The validation dataset contains 15 videos; 5 sets which contain 3 videos at each set (Figure \ref{multiple_camera}) and each video consists of 300 frame images. This is the same as the test set used in the pilot study. The test dataset is 500 sets of videos of the 2020 AI Grand Challenge\footnote{AI Grand Challenge link - https://www.ai-challenge.or.kr/}.  However, the data was not open to the challengers. We only received the score of our model on the Test set.

\textbf{Parameter}. We set parameters for object detecting and tracking for better performance like Table \ref{2020_param_setting}. One of the tips is that we tuned the tracklet buffer to prevent fragmentation caused by the large motions of cameras. It was our strategy to extract proper IDs when tracking, then, to embed and group them to obtain unique IDs. Euclidean distance was used as the metric for person ID association, which may be naive yet simple and easy to analyze.

\begin{table}[ht]
\begin{center}
\begin{sc}
\begin{tabular}{c|c} \hline
frames to process per second & 270 / 300 \\\hline
frame buffer for tracking & 180 \\\hline
detection threshold  & 0.3 \\\hline
confidence threshold & 0.6 \\\hline
non maximum suppression threshold & 0.4 \\\hline
\end{tabular}
\end{sc}
\end{center}
\caption{(Study-1) Hyper parameter setting}
\label{2020_param_setting}
\end{table}

\textbf{Implementation detail}. 
Here, the resource constraints for Study-2 are illustrated in Table \ref{2020_experiment_environments}. Two Tesla V100 GPU's are provided in the inference environment. The task has to be completed within a limited condition of 6 hours. 500 samples of test data need to be processed. 20.8 frames per second were required to process tasks of 300 frame jpeg images within the limited time.

\begin{table}[ht]
\begin{center}
\begin{sc}
\begin{tabular}{c|c} \hline
GPU & Tesla V100  \\\hline
test dataset (set) & 500 \\\hline
inference time (hour) & 6 \\\hline
\end{tabular}
\end{sc}
\end{center}
\caption{(Study-1) Inference conditions}
\label{2020_experiment_environments}
\end{table}

\subsection{Results}

Comparison with the state-of-the-art methods is difficult. So, we just count the unique number after person re-identification with the small datasets. The accuracy of FairMOT is 85.44\% and our model is 85.71\%, which is a slight improvement in our quantitative result. It is an observation to improve counting the unique persons through re-identification in the second step. In the test dataset, the score of the combination of FairMOT and OSNet was much smaller. In the results calculated based on L2-norm error, the baseline was 48.1, while the proposed model combination was 34.9, which is a great reduction of error by a margin of 27.4\%. The use of fast trackers and light re-identification models enabled processing of 27 frames per second (FPS) close to real-time and reduced the final error to 28.7 as shown in Table \ref{2020_quan_res}. Our qualitative analysis further proves the efficacy of the combination of FairMOT and OSNet, as shown in Figure \ref{2020_qual_res}. In contrast, the baseline model is shown to have difficulty in recognizing which person is which.

\begin{table}[ht]
\begin{center}
\begin{sc}
\begin{tabular}{|c|C{28mm}|C{28mm}|C{28mm}|} 
\hline
Quantitative Result & FairMot & \multicolumn{2}{c|}{FairMot+OSNet (ours)} \\
\hline
validation dataset (\%) & 85.44 & 85.71 & - \\
\hline
test set (L2-norm) & 48.14 & 34.93 & 28.7 \\
\hline
fps & 15 & 15 & 27 \\
\hline
frame buffer & 150 & 150 & 180 \\
\hline
detection threshold & 0.4 & 0.4 & 0.3 \\
\hline
\end{tabular}
\end{sc}
\end{center}
\caption{(Study-1) FairMOT vs. FairMOT + OSNet}
\label{2020_quan_res}
\end{table}

\subsection{Discussion}
There are many problems with multi-camera tracking. In terms of object detection, the model may have difficulty in detecting a person on a bike, or a person behind the glass door. In the case of object tracking, ID switching or fragmentation occurred, and was vulnerable to the global motions of the camera. The ID switching problem was resolved by increasing the detection threshold and re-identifying the person with camera tracklet association at a later stage. Most of the tracking errors were solved by tracklet association.

In this section, we can improve re-identifying the sample person in multiple cameras by combining these two fast tracker and light re-identification, whose strength is in multi-scales and multi-views. It requires a smaller cost and shows a more suitable configuration for objects with less global motion and more local movement.

\begin{figure}[ht]
\centering
\includegraphics[width=0.98\textwidth]{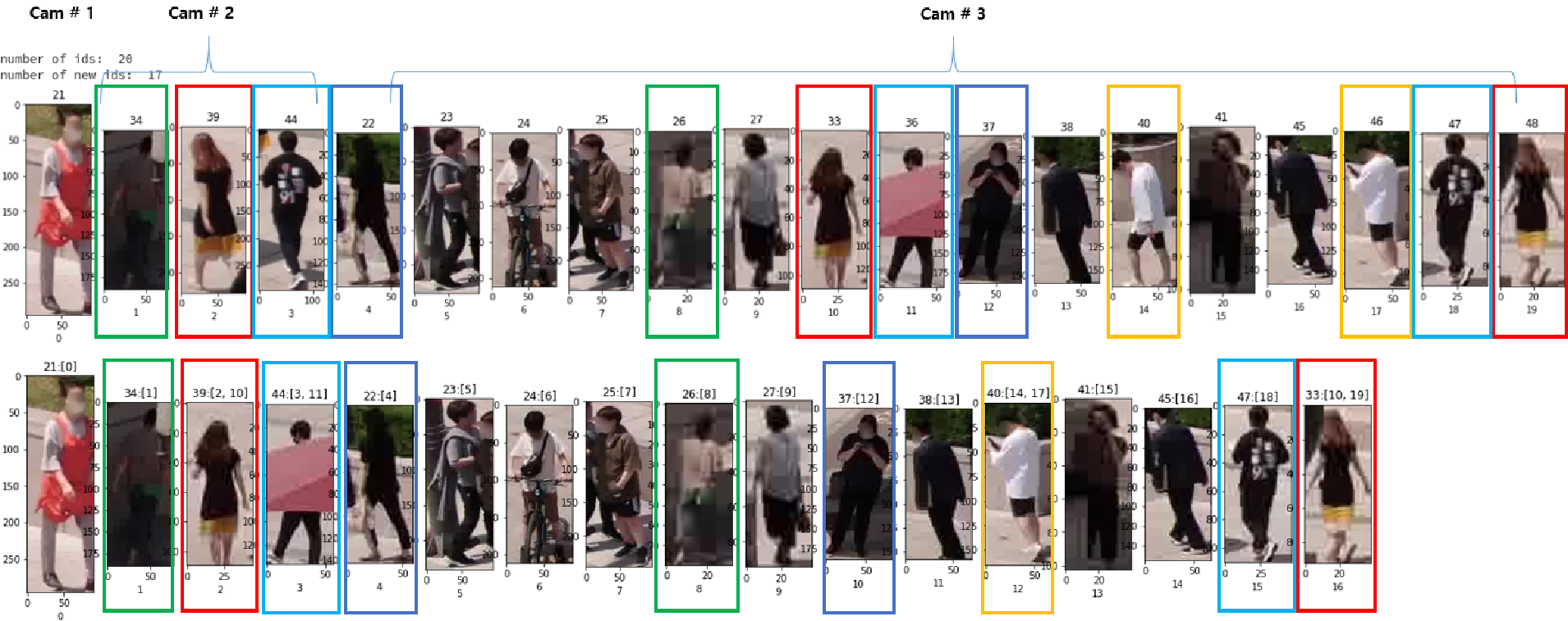}
\\
\includegraphics[width=0.98\textwidth]{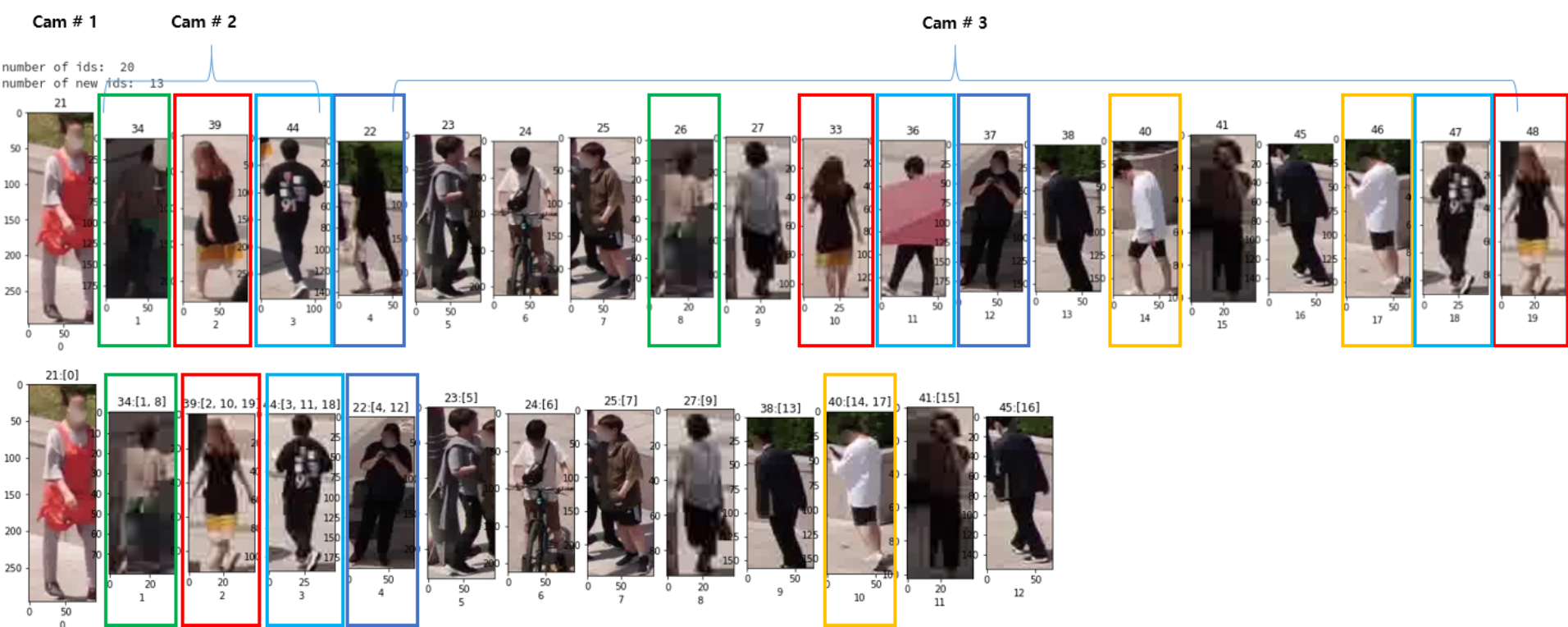}
\caption{(top) FairMOT only. (bottom) FairMOT + OSNet.}
\label{2020_qual_res}
\end{figure}

In Figure \ref{2020_qual_res}, re-ID results of the baseline model (top) and our model (bottom) is demonstrated. The top row of each model is the cropped images extracted from different cameras, and the bottom row contains the re-ID results. The ground truth labels, i.e., the ground truth ID, are represented in different colored boxes. In the figure, we can see that the baseline model confuses the same person for different IDs. That is, the re-ID results contain image snippets with the same colored box, which means the model could not assign the same ID to the same person extracted from different cameras. However, our model at the bottom could successfully categorize the cropped images. We can observe that only one image is assigned per box color. The Euclidean distance metric was used to merge outputs from different cameras. The voting method showed more accurate performance in the validation dataset. However, its performance tended to deteriorate in the test set, which implies its poor generalizability. Also, the limited inference time is one of our burdens. The use of the fast one-shot object tracker and the light-weighted Re-ID helped secure efficiency.

\section{Study-2: The Unique Persons in Moving Cameras}

\subsection{Problem Definition}
The purpose of this experiment is to discover the person who has the same ID in multiple cameras, and to determine whether each person is moving or not. While multiple-view videos are used in the previous study, one of the differences in our new dataset is that the situation of the video changes depending on time. In the video, the target objects, mannequins, are moving a part of their body like arms or hands, while their locations are stationary.

The drone flies in the same routine three times. This is like a simulation of rescuing people in a kind of disaster situation. As a restriction, rescuers such as firefighters, medical staff should be excluded from the count, as only the number of people to be rescued would be in need in real-world applications.

First of all, performance detection should be applied to tracking under irregularly moving camera conditions. Depending on the movement of a camera, object tracker cannot catch objects' trajectory caused by global motion and ID fragmentation. The time limit was the next problem. We tried to solve the inference issues through implementation techniques.

\subsection{Architecture Overview}

The results of Study-1 show the importance of matching type and the suitable combination of object tracker and re-identification. However, all of the objects are not that have small global motion. 

We assume that the detection performance is more important than the case of Study-1. CenterNet is a fast object detector, but its performance potential can be degraded because it does not rely on prior knowledge. It may be one method to apply other alternatives such as YOLOv5 for more accurate performance. In order to find a more suitable combination for the current conditions, we apply the two-stage detection method that achieves higher accuracy.

\begin{figure}[ht]
\centering
    \includegraphics[width=1.0\textwidth]{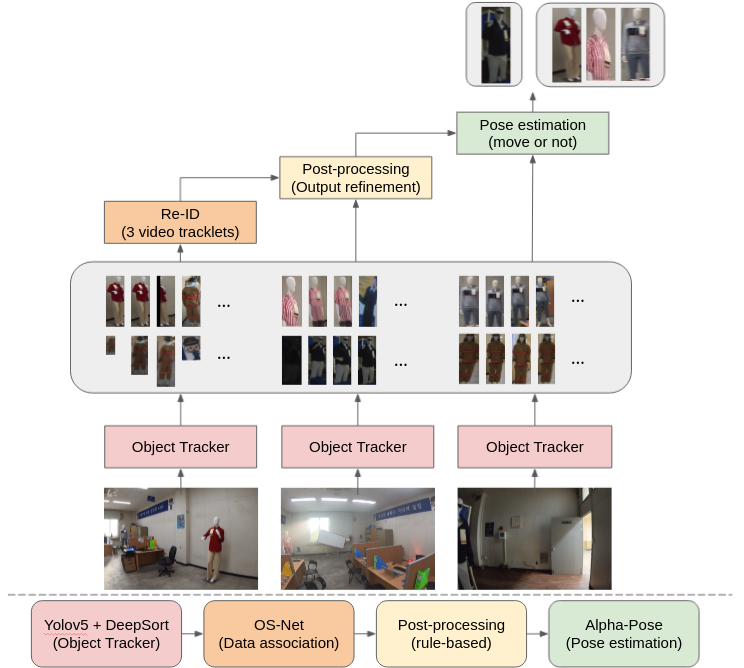}
    \caption{Pipeline in a moving-camera system.}
    \label{2021_pipeline}
\end{figure}

Figure \ref{2021_pipeline} shows that we replace FairMOT with DeepSort \cite{yu2018deep} that adopts YOLOv5 to reinforce detection recall attributes. The re-ID part almost remains in the same form, and the new pipeline contains an output refinement block to remove false positives which arise in the detection part after the re-ID stage. Plus, Pose estimation is the new part of the architecture to detect the objects' local movements. As pose estimation falls outside of our range of study, its further discussion is omitted in this work.

\subsection{Experiment Setup}

\begin{figure}[ht]
\centering
    \includegraphics[width=1.0\textwidth]{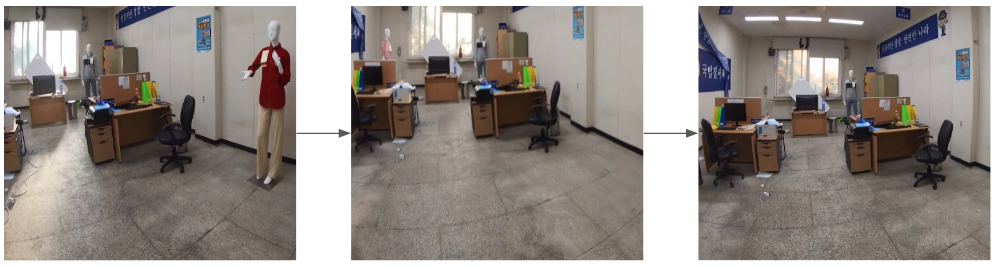}
    \caption{Validation dataset by a moving camera.}
    \label{2021_disaster}
\end{figure}

\textbf{Dataset}. MS-COCO is the training dataset for object detection and the pre-trained data from market1501 was used in re-identification. The validation dataset contains 3 videos; 1 set contains 3 videos. Each video is about 5 minutes long. It is more or less 7000 frame images. 
Most of the objects are mannequins and some are moving a part of the body, while others are not, as shown in Figure \ref{2021_disaster}. The test dataset consists of 5 sets of videos of the 2021 AI Grand Challenge\footnote{AI Grand Challenge link - https://www.ai-challenge.or.kr/}. However, it is not open to the challengers. We only received our model's score on the test dataset.

\textbf{Parameter}. We set many parameters in a heuristic way for object detecting and tracking for better performance as shown in Table \ref{2021_param_setting}. Max age and min confidence are important factors to track objects from avoiding the huge motion of a camera. Some parameters such as the nearest neighbor budget, maximum distance, confidence threshold, and maximum distance for handling object detection and tracking performance have been adjusted in detail. The Euclidean distance for person association still works well and is simple to analyze results. Pose threshold and variance are new hyper parameters to clarify the movement of objects of interest. One more interesting parameter is Nearest Neighbour Similarity, which is generally used as cosine distance. However, we applied the Euclidean distance as appearance matching after an experimental review.

\begin{table}[ht]
\begin{center}
\begin{sc}
\begin{tabular}{c|c} \hline
max age & 250 \\\hline
nn budget & 100 \\\hline
nn similarity & the Euclidean distance \\\hline
min confidence  & 0.65 \\\hline
max distance & 0.05 \\\hline
conf threshold & 0.25 \\\hline
min conf threshold & 0.65 \\\hline
max distance & 0.05 \\\hline
min object size & width > 60 and height > 50 \\\hline
pose threshold & 0.25 \\\hline
pose variance & 0.1 \\\hline
\end{tabular}
\end{sc}
\end{center}
\caption{(Study-2) Hyper parameter setting}
\label{2021_param_setting}
\end{table}

\textbf{Implementation detail}.
One Tesla P40 GPU and 4-core CPU are only allowed in the inference environment, which has 62.73\% lower performance than the resource provided in the previous study. The task has to be completed within a limited condition of around 23 minutes in Tesla V100, and 35 minutes in P40. Only 5 sets of test data need to be processed, each with 5 minutes in length. 63 frames per second were required to process tasks within a limited time in Tesla P40. When comparing with Study-1 under the same conditions, 101 frames per second is needed to complete all tasks, which requires about 5 times inference speed in Table \ref{2021_experiment_environments}. To overcome this time budget within the limited test environment, we apply multi-threading and use only a quarter of all the frames. 

\begin{table}[ht]
\begin{center}
\begin{sc}
\begin{tabular}{c|c|c|c} 
\hline
comparison & Study-1 & Study-2 & ratio \\
\hline
H/W Performance & Tesla V100 & Tesla P40 & $\approx{0.63}$ \\
\hline
test dataset (set) & 500 & 5 & - \\
num of video (EA/set) & 3 & 3 & - \\
num of frame (Frame/EA) & 300  & 9000 & - \\
task (frame) & 450k & 135k & 0.3 \\
\hline
(V100) inference time (min) & 360 & 23 & 0.06  \\
\hline
(V100)  target run-time (fps) & 20.83 & 100.99 & $\approx{4.85}$ \\ 
\hline
\end{tabular}
\end{sc}
\end{center}
\caption{(Study-2) Inference conditions}
\label{2021_experiment_environments}
\end{table}

\subsection{Results}

Although DeepSORT only processes a quarter of all frames due to hardware and time limitations, Table \ref{2021_quan_res} shows that DeepSORT (42.9\%) outperforms FairMOT (71.4\%) in terms of recall in fast-moving drone cameras. It is meaningful to see the recall and f1 score. It is also important to accurately recognize fewer people in disaster situations, but it is more important to identify more rescue targets.

\begin{table}[ht]
\begin{center}
\begin{sc}
\begin{tabular}{|c|c|c|} 
\hline
Quantitative Result & FairMOT + OSNet & DeepSORT + OSNet \\
\hline
Accuracy (\%) & 42.9 & 62.5 \\
\hline
Recall (\%) & 42.9 & 71.4 \\
\hline
F1 Score (\%) & 60 & 76.9 \\
\hline
\end{tabular}
\end{sc}
\end{center}
\caption{(Study-2) Validation dataset. FairMOT + OSNet. vs. YOLOv5-DeepSORT + OSNet.}
\label{2021_quan_res}
\end{table}

\begin{figure}[ht]
\centering
    \subfigure[Tracks after object tracking.]{\includegraphics[width=1.0\textwidth]{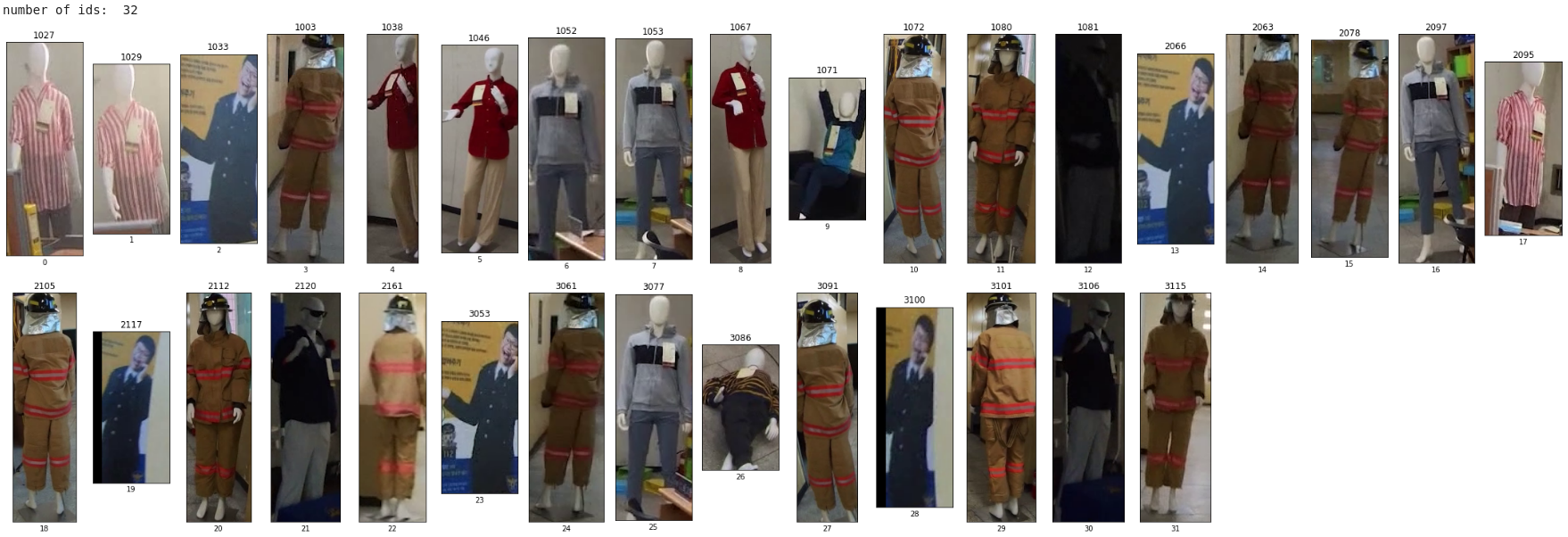}}
    \vspace{0.01cm}
    \hrule
    \vspace{0.01cm}
    \subfigure[Tracklets association.]{\includegraphics[width=0.8\textwidth]{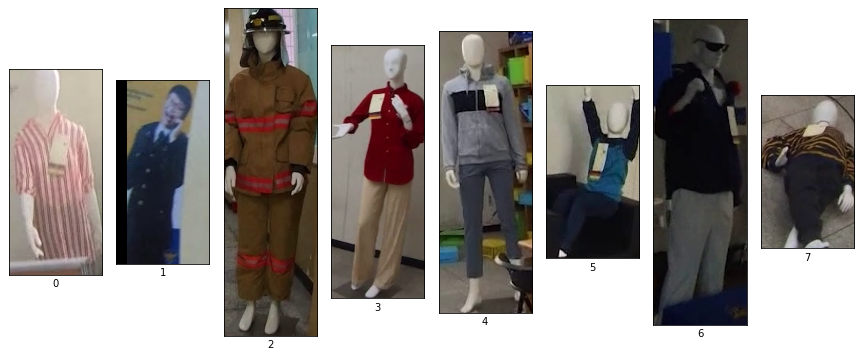}}
    \vspace{0.01cm}
    \hrule
    \vspace{0.01cm}
    \subfigure[Final candidates to count after post-processing.]{\includegraphics[width=0.8\textwidth]{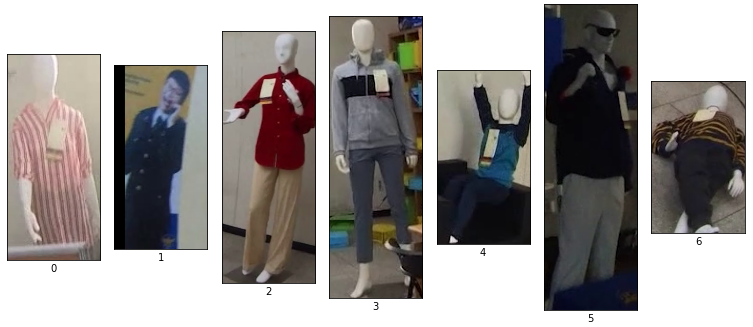}}
    \caption{YOLOv5-DeepSORT + OSNet. Counting the unique IDs in moving cameras.}
    \label{2021_qualitative_result}
\end{figure}

Clarifying non-person in re-ID and Anomaly detection like Spade can be supplemented to remove false positives like posters and non-person objects in many heuristic ways as like \ref{2021_qualitative_result}. Moreover, pose estimation such as Alpha-pose can predict the person with movement or not. 

\clearpage

\subsection{Discussion}
In this section, we aim to understand the suitable combination of object tracker and re-identification in various situations. FairMOT with OSNet was adequate for dense objects with less global motion and more local movements as in the previous study. A two-stage object tracker was the better choice for fast-moving camera videos. We increased the recall performance of the detector and removed false positives with the output refinement block. It was a better strategy than concentration with precision.

The number of false positives and false negatives interfere with person re-identification. Our heuristic methods have many weaknesses, such as the limitation of distinguishing non-person objects by threshold. Such problems can be resolved with deep learning-based methods with more data in the future. Adding a `poster' class when training the detection model can be one method to further remove false positives.

The inference time was another challenge under limited resources. When it comes to resource comparison between Study-1 and Study-2, CPU and GPU conditions are under 62.73\% than before, the processing time is allowed less than 9.87\%. Overall, we had to improve execution tasks 4.85 times faster than the previous study, although 30\% of the tasks need to be processed. In system implementation, only 25\% of video frames are processed and multi-threading methods are another good choice to reduce the latency to 74.38\%. As a result, these implementation techniques make it possible to handle all tasks within the time limit. It will be an important option for us to make the model efficient.

\chapter{Conclusion}

We propose two approaches that do not only find appropriate combinations of detectors, trackers, and Re-ID models but also resolve actual implementation difficulties.

Study-1 shows a fast and light multi-camera system. It is a simple approach that incorporates the advantages of both models. FairMOT is a simple but lighter and faster object tracker. OSNet also has a lightweight CNN architecture that is capable of learning omni-scale feature representations. Accordingly, our model is light and fast, maintaining the advantages of FairMOT and OSNet. Moreover, we showed that it can perform better than using FairMot alone. The quantitative result is improved slightly and we can check better results with unique person ID matching of qualitative evaluations. In order to count the same person for multi-views, it is possible to improve the tracking performance of the object using a suitable object tracker and re-ID. 

YOLOv5 is a fast one-stage object detector and focuses on performance. Moreover, many recently proposed powerful object trackers are built on DeepSORT. We combined and implemented the model with a focus on detection performance regarding the difficulties of irregular movement of the camera and disturbances in the background. It alleviates the ID-switching problem caused by global motion and fragmentation. 
Also, by increasing detection performance in moving cameras, the object tracker is able to capture more objects and achieve better tracking performance by removing false positives with the output refinement block.

In the future, we will investigate better metrics because it was not easy for us to use traditional methodologies. The multi-camera system will be improved through a combination of various proper object trackers and re-ID solutions considering Study-1 and Study-2. Plus, further improvements in the output refinement block may lead to a greater impact on overall performance. Also, proposing a one-stage Online Multi-Camera Tracking model is one of our future works. It is more challenging, but makes Multi-Camera-Multi-Object-Tracking lighter and faster in terms of efficiency.

\end{document}